\documentclass[letterpaper,10pt,conference]{IEEEtran}

\usepackage{amsmath,amssymb,amsfonts}
\usepackage{graphicx}
\usepackage{booktabs}
\usepackage{cite}
\usepackage{hyperref}
\usepackage{xcolor}
\usepackage{enumitem}
\usepackage[caption=false,font=footnotesize]{subfig}
\usepackage{balance}
\usepackage{multirow}

\hypersetup{
  colorlinks=true,
  linkcolor=black,
  citecolor=black,
  urlcolor=black
}

\newcommand{\paperfigure}[2]{%
  \IfFileExists{#1}{%
    \includegraphics[width=\columnwidth]{#1}%
  }{%
    \fbox{%
      \parbox[c][1.25in][c]{0.95\columnwidth}{%
        \centering
        \footnotesize #2\\[2pt]
        \scriptsize Insert file: \texttt{#1}%
      }%
    }%
  }%
}

\title{Bootstrap Perception Under Hardware Depth Failure\\for Indoor Robot Navigation}

\author{%
  \IEEEauthorblockN{Nishant Pushparaju, Vivek Mattam, Aliasghar Arab$^{*}$}%
  \thanks{$^{*}$Corresponding author.
  N.~Pushparaju and V.~Mattam are with the Department of Mechanical and Aerospace Engineering, New York University, New York, NY 11201, USA.
  A.~Arab is the corresponding author and is affiliated with both the City College of New York and New York University (Department of Mechanical and Aerospace Engineering).
  Emails: {\tt\small \{np3129, vm2677\}@nyu.edu}; A.~Arab: {\tt\small aarab@ccny.cuny.edu}, {\tt\small aliasghar.arab@nyu.edu}.}%
}

\begin{document}
\maketitle

\begin{abstract}
We present a bootstrap perception system for indoor robot navigation under
hardware depth failure. In our corridor data, the time-of-flight camera loses up
to 78\% of its depth pixels on reflective surfaces, yet a 2D LiDAR alone
cannot sense obstacles above its scan plane. Our system exploits a
self-referential property of this failure: the sensor's surviving valid
pixels calibrate learned monocular depth to metric scale, so the system
fills its own gaps without external data. The architecture forms a failure-aware sensing hierarchy,
conservative when sensors work and filling in when they fail: LiDAR remains
the geometric anchor, hardware depth is kept where valid, and learned depth
enters only where needed. In corridor and dynamic pedestrian evaluations,
selective fusion increases costmap obstacle coverage by 55--110\% over
LiDAR alone. A compact distilled student runs at 218\,FPS on a Jetson Orin
Nano and achieves 9/10 navigation success with zero collisions in
closed-loop simulation, matching the ground-truth depth baseline at a
fraction of the foundation model's cost.
\end{abstract}

\begin{IEEEkeywords}
depth estimation, sensor fusion, knowledge distillation, costmap, indoor navigation
\end{IEEEkeywords}

\section{Introduction}

Indoor robots are built around range sensors that are reliable until the scene
violates their assumptions. A 2D LiDAR captures stable angular structure, but
only on a single plane. Time-of-flight (ToF) cameras give direct metric depth;
however, polished floors, glass doors, and the reflective surfaces common
in warehouses, manufacturing, and logistics cause them to return invalid measurements. While a 3D
LiDAR would solve the vertical coverage problem, its cost, weight, and power
exceed the budget of compact indoor platforms; the 2D LiDAR plus camera
combination is the dominant configuration in this deployment class~\cite{nav2,turtlebot3}. In the
corridor data used in this paper, an Orbbec Femto Bolt returns invalid depth
($d{=}0$ or $d{>}5$\,m) on a per-pixel mean of 45\% across 12\,384 frames
from a 410\,s corridor recording, computed by classifying every pixel with zero
return or range beyond the sensor's 5\,m indoor validity cutoff as invalid.
In a separate high-invalidity segment of the same corridor, dominated by
polished floors and glass doors, invalidity rises to a mean of 78\% across
459 frames. The sensor
remains active, yet the geometry available to the navigation stack becomes
incomplete. In our tested deployment setting, this was a persistent failure
mode rather than an occasional outlier.

We address this with a bootstrap perception system. We call it
\emph{bootstrap} because the failing sensor's surviving valid pixels provide
the calibration signal that converts learned relative depth into metric depth,
so the system fills its own gaps instead of degrading to LiDAR alone.
LiDAR remains the geometric anchor; hardware depth is used where valid;
learned monocular depth fills only the invalid regions; and semantics are
added only where geometry alone is insufficient. The stack is conservative
when sensors work and fills in when they do not.
To our knowledge, using a degraded depth sensor's own surviving pixels to
calibrate learned depth at runtime --- and filling that sensor's dead pixels
with the calibrated output, without offline data or external ground
truth --- has not been previously demonstrated.

We evaluate in three settings. First, we test whether learned depth adds
obstacle coverage in a university corridor where polished floors and glass
cause severe active-depth failure (\S\ref{sec:r1}). Second, we test whether
learned depth adds above-plane obstacle evidence on
LILocBench~\cite{lilocbench}, a public long-term indoor localization
benchmark with moving pedestrians, used here for costmap evaluation
(\S\ref{sec:r2}). Third, we ask whether a compact distilled student can
replace the foundation model for embedded navigation, and bound its near-range
safety (\S\ref{sec:r3}). Platform deployment constraints are reported first.

The contributions are:
\begin{enumerate}[nosep,leftmargin=*]
  \item A bootstrap perception architecture where LiDAR, hardware depth,
  learned monocular depth, and optional semantics enter the local costmap in a
  deliberate hierarchy, not as independent modules.
  \item A systems evaluation: in corridor replay, LiDAR+depth increases
  occupied obstacle cells by 55\%; on dynamic LILocBench, by 110\% over
  LiDAR alone. A corridor-specialized student matches the ground-truth-depth
  baseline on the reported Gazebo corridor metrics, achieving 9/10 success
  with zero collisions.
  \item A distillation ablation from teacher to compact student, showing
  that corridor specialization is viable but requires accepting
  generalization and near-range safety tradeoffs.
\end{enumerate}

All real-world data was collected by teleoperation. The evaluation combines
offline replay on recorded sensor data with closed-loop Gazebo simulation.

\begin{figure}[!t]
\centering
\paperfigure{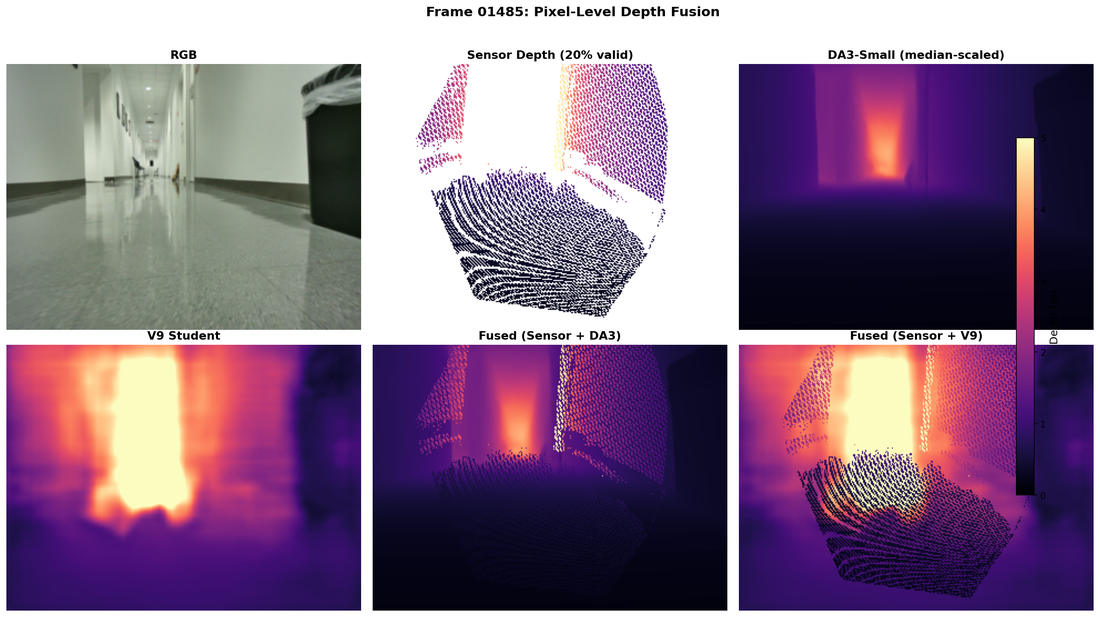}
{Pixel-level depth fusion at worst-case corridor failure (20\% valid pixels).}
\caption{Depth fusion under worst-case ToF failure (20\% valid pixels). The
hardware sensor retains only a sparse stippled pattern; DA3-Small supplies
dense depth estimates in the regions where the hardware sensor fails. Even at 80\% pixel failure, the fused output
combines surviving hardware depth with learned depth to produce a dense
depth estimate for costmap construction.}
\label{fig:depth-fusion-worstcase}
\end{figure}

\section{Related Work}

\textbf{Monocular depth estimation.}
Recent monocular depth models have made dense geometry available from a single
RGB frame. MiDaS~\cite{midas} established strong mixed-dataset
transfer. Depth Anything~\cite{da1,da2,da3} scaled that
idea with larger encoders and larger unlabeled corpora. On benchmark datasets,
supervised methods such as AdaBins~\cite{adabins},
BTS~\cite{bts}, and PixelFormer~\cite{pixelformer} remain
strong references. Dense monocular geometry is now achievable
across many scene types, but a deployment question remains open: once an
active-depth sensor becomes locally unreliable, how should learned depth
enter a navigation stack, and what evidence shows that it improves the
geometry available to the planner?

\textbf{Depth in navigation systems.}
Depth has long been used for mapping, traversability, and embodied navigation.
Gupta~et~al.~\cite{gupta2017cognitive} build ego-centric metric maps from
predicted depth for visual navigation; Chaplot~et~al.~\cite{chaplot2020}
use depth to construct top-down occupancy maps for active exploration;
ViPlanner~\cite{viplanner} learns traversability costs from semantic depth.
Predicted depth can clearly support spatial reasoning, but the
cited systems all assume either ground-truth depth or a healthy active
sensor. How learned depth changes a real local costmap when the active
depth sensor itself is degraded is less well documented.

The closest prior work occupies adjacent but distinct problem settings.
LiDAR-guided depth completion~\cite{tao2022} densifies sparse 3D LiDAR
using camera guidance, assuming both sensors are healthy; it does not model
per-pixel ToF invalidity or substitute a learned signal selectively.
Transparent-obstacle detection~\cite{topgn2024} identifies glass and
reflective surfaces via LiDAR intensity profiles but does not substitute a
learned depth estimate in the navigation costmap.
Recent monocular-depth integration into Nav2~\cite{depthav2} uses camera
depth as a full LiDAR replacement, not a selective fallback.
Prior work either replaces the primary range source entirely, assumes both
sensors are healthy, or targets a different obstacle modality. The policy
studied here (keeping LiDAR as the SLAM anchor, using hardware ToF depth
where valid, and selectively inserting learned monocular depth only in
per-pixel-invalid regions of the local costmap) is not directly
addressed in any of these lines of work.
None of the cited approaches use the degraded sensor's own surviving pixels
to calibrate and fill its dead regions at runtime --- the self-calibrating
mechanism central to this work.

\textbf{Efficient edge perception and distillation.}
EfficientViT~\cite{efficientvit} and
EfficientViT-SAM~\cite{efficientvitsam} show that dense prediction
and mask refinement can be run within embedded budgets. Kendall-style multi-task
weighting~\cite{kendall2018} and berHu depth losses~\cite{berhu} remain
practical tools for compact student training. Compact runtime perception is
therefore plausible, but the question of when a specialist should replace a
stronger general model is less studied. The student study in this paper addresses that deployment
boundary directly by comparing general indoor performance, corridor
specialization, and near-range safety instead of only reporting standalone
benchmark gains.

\section{System Overview}

\subsection{Deployment Setting}

The system is deployed on a compact indoor robot platform equipped with an
Orbbec Femto Bolt (1280${\times}$720 RGB, 640${\times}$576 ToF depth),
a 2D LiDAR, and a Jetson Orin Nano 8\,GB for inference. ROS2 Humble with
Nav2~\cite{nav2} and SLAM Toolbox~\cite{slamtoolbox} provides navigation
and localization.

The sensing stack has four layers. LiDAR supplies stable planar geometry and
remains the primary spatial reference. Hardware depth gives direct metric
measurements when the observed surface is measurable, but when it becomes
locally unreliable, monocular depth fills the gap with dense estimates.
Where geometry alone is not sufficient, semantics add class-dependent
obstacle refinement.

Heavy teacher models remain off-board: DA3-Large~\cite{da3} provides depth
supervision and SAM2-Large~\cite{sam2} generates instance masks that are
mapped to the six semantic classes (floor, wall, person, furniture, glass,
other) via YOLOv8~\cite{yolov8} detection labels. Both run on server-side
GPUs for offline label generation only. Two
deployable learned-depth paths share the same sensing role on the robot:
DA3-Small as the general-purpose foundation model, and a compact distilled
student for specialized corridor deployment. The paper evaluates both;
the strongest general-purpose costmap evidence is from DA3-Small, while
the student is validated through offline metrics and closed-loop simulation.

\subsection{Failure-Aware Sensing Hierarchy}

\begin{figure*}[!t]
\centering
\includegraphics[width=0.96\textwidth]{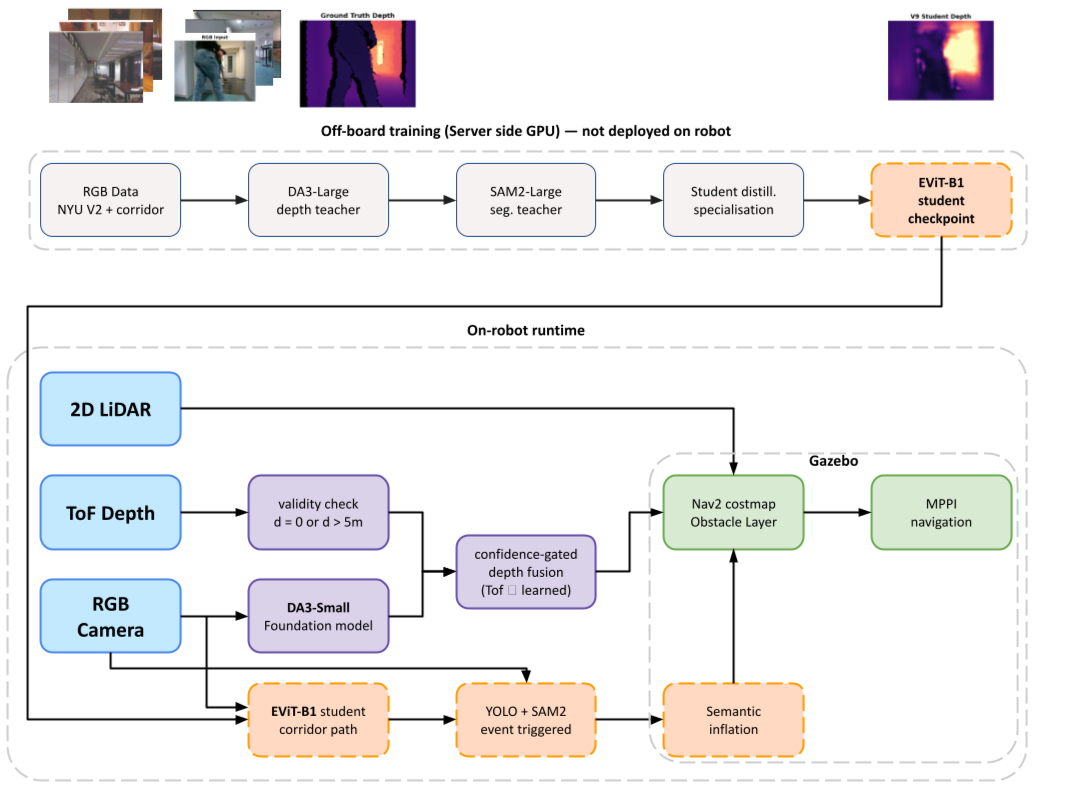}
\caption{System architecture. LiDAR remains the primary geometric reference,
hardware depth is used where valid, learned monocular depth replaces unreliable
active-depth regions, and optional semantics refine obstacle footprint only
when needed. Heavy teachers remain off-board. Dashed orange paths are optional.}
\label{fig:system-architecture}
\end{figure*}

The system is organized as a sensing hierarchy. When the standard sensors are reliable, the stack uses them
conservatively. When those sensors become incomplete, the system fills the
missing regions with learned depth without discarding the stable parts of the
sensing stack. This supports both failure modes studied in the paper:
reflective-surface degradation in corridor sensing and scan-plane
incompleteness in dynamic scenes. The data flow is summarized in
Fig.~\ref{fig:system-architecture}.

Operationally, the system performs five steps. It first filters active-depth
measurements for validity. It then calibrates monocular depth into metric
form. The resulting depth map is reprojected into 3D and filtered to the
collision-relevant height band before publication as obstacle evidence. LiDAR
and learned depth are fused at the costmap level. Finally, semantic cues may
refine obstacle inflation radii, assigning a wider safety margin around
moving pedestrians than around static walls, or explicitly marking glass
surfaces that are transparent to LiDAR and invisible to fixed inflation.

\subsection{Embedded Constraint}

Embedded compute constrains the design. The Jetson
measurements (Table~\ref{tab:perception-stack}) show that DA3-Small,
YOLOv8~\cite{yolov8}, and event-triggered segmentation refinement fit within
roughly 6.5\,GB. This is sufficient for runtime
deployment, but leaves little headroom once the rest of the autonomy stack is
active. That constraint drives two design choices: selective use of semantics
(triggered, not always-on) and exploration of the compact student path
for narrow environments where domain-specific specialization may justify a lower
runtime cost.

\subsection{Depth Calibration, Fusion, and Costmap Construction}

Let $\hat{d}(u,v)$ denote the monocular depth predicted at pixel $(u,v)$,
$d_{\mathrm{tof}}(u,v)$ the active-depth measurement, and
$c(u,v) \in [0,1]$ the per-pixel confidence reported by the Femto Bolt
sensor (derived from the IR return amplitude; low values indicate
specular reflection or absorption).
Both DA3-Small and the student output relative depth; metric units are
recovered by a per-frame scale factor computed via median alignment against
the live ToF depth:
\begin{equation}
d(u,v) = s \cdot \hat{d}(u,v), \quad
s = \operatorname{median}_{\substack{c(u,v)\ge\tau,\\ \hat{d}(u,v)>\epsilon}}\!\left(\frac{d_{\mathrm{tof}}(u,v)}{\hat{d}(u,v)}\right),
\label{eq:scale}
\end{equation}
with confidence threshold $\tau{=}0.5$ and $\epsilon{=}10^{-3}$\,m to
exclude degenerate predictions.
This is the bootstrap: the same sensor whose failure motivates learned-depth
substitution supplies the valid pixels that anchor it to metric scale.
Even when 78\% of pixels are invalid, the remaining 22\% (roughly
81\,000 pixels on a 640${\times}$576 frame) suffice for a stable median.
In the corridor data the per-frame scale factor $s$ has a coefficient of
variation below 8\% across the 459-frame worst-case segment; the median is
robust to outlier ToF returns and to the spatial distribution of surviving
pixels. Because surviving pixels span the sensor's full observed depth
range (floor, wall, and object returns from 0.05 to 5\,m), the median is
not dominated by a single depth band. The calibration requires no offline
data and adapts on every frame.
Below ${\sim}5\%$ valid coverage the scale estimate becomes unreliable;
this boundary was not reached in the tested data.

The runtime fusion rule is
\begin{equation}
d_{\mathrm{fused}}(u,v)=
\begin{cases}
d_{\mathrm{tof}}(u,v), & c(u,v)\ge \tau \;\text{and}\; d_{\min}<d_{\mathrm{tof}}<d_{\max} \\
d(u,v), & \text{otherwise},
\end{cases}
\label{eq:fusion}
\end{equation}
with $\tau{=}0.5$, $d_{\min}{=}0.05$\,m, $d_{\max}{=}5.0$\,m.
Hardware depth is kept where valid; learned depth fills the rest.

Fused depth values are back-projected into 3D using the camera intrinsic matrix
$\mathbf{K}$ and transformed into the robot base frame via the extrinsic
$\mathbf{T}^{B}_{C}$. Only points within the height band
$[h_{\min}, h_{\max}] = [0.05, 2.0]$\,m enter the costmap. The resulting point set is published
as a ROS \texttt{PointCloud2} and fused with LiDAR-derived obstacle evidence
in the local costmap.

\subsection{Compact Student Training Objective}

The compact student predicts both depth and segmentation from RGB. Training
uses a supervision rule that mirrors the runtime fusion:
\begin{equation}
d_{\mathrm{target}}(u,v)=
\begin{cases}
d_{\mathrm{tof}}(u,v), & c(u,v)\ge \tau \\
d_{\mathrm{teach}}(u,v), & c(u,v)<\tau,
\end{cases}
\label{eq:hybrid-target}
\end{equation}
where $d_{\mathrm{teach}}$ is the teacher depth (DA3-Large). The range check
($d_{\min}$, $d_{\max}$) from Eq.~\ref{eq:fusion} is applied during
target construction but omitted from the notation for brevity. In NYU
training, all ground-truth pixels satisfy the confidence threshold, so the
target reduces to direct supervision; in corridor settings the student
inherits teacher supervision exactly where the active sensor is least
reliable.

The loss combines three terms: a berHu depth loss~\cite{berhu}
$\mathcal{L}_{d}$ (L1-like on small errors, quadratic on large errors; more
stable than SILog in the compact-model runs), a segmentation cross-entropy
$\mathcal{L}_{s}$, and an edge-aware smoothness regularizer
$\mathcal{L}_{\mathrm{smooth}}$ that preserves sharp depth boundaries at
glass, wall, and furniture transitions. Depth and segmentation weights are set
by clamped Kendall uncertainty weighting~\cite{kendall2018} so that neither
task dominates:
\begin{equation}
\mathcal{L}
=
\tfrac{1}{2}e^{-\log \sigma_d^2}\mathcal{L}_{d}
+
e^{-\log \sigma_s^2}\mathcal{L}_{s}
+
\tfrac{1}{2}\log \sigma_d^2
+
\tfrac{1}{2}\log \sigma_s^2
+
\lambda_e \mathcal{L}_{\mathrm{smooth}},
\label{eq:kendall-loss}
\end{equation}
with $\log \sigma_d^2$ and $\log \sigma_s^2$ clamped to $[-2, 2]$ during
optimization to prevent task collapse, and $\lambda_e{=}0.1$.

The student uses an EfficientViT-B1 encoder (5.31M parameters) with two
lightweight decoders sharing skip connections at 1/4, 1/8, and 1/16 resolution.
Each decoder applies three transposed-convolution stages with skip fusion,
producing depth (1-channel, ReLU) and segmentation (6-class: floor, wall,
person, furniture, glass, other) at $240{\times}320$, bilinearly upsampled to
input size. Training uses AdamW ($\mathrm{lr}{=}10^{-3}$, encoder
$0.1\times$, weight decay $10^{-4}$) with cosine annealing for 200 epochs on
NYU (batch 16) or 50 epochs for corridor fine-tuning (batch 128). The
encoder is frozen for the first 5 epochs. Gradient norms are clipped at 1.0.
Input is $240{\times}320$ RGB with ImageNet normalization. Deployment
augmentation (V5 onward) adds random rotation ($\pm 10^\circ$), scale jitter
(0.85--1.15), Gaussian blur, JPEG compression artifacts, and random cutout to
close the domain gap between NYU training and on-robot inference.

\section{Experiments and Results}

We evaluate in three settings preceded by platform deployment
(\S\ref{sec:platform}): corridor costmap recovery under active-depth failure
(\S\ref{sec:r1}), dynamic obstacle recovery on a public pedestrian benchmark
(\S\ref{sec:r2}), and compact student deployment for embedded navigation
(\S\ref{sec:r3}). All real-world data was collected by teleoperation and
replayed offline.

\emph{Datasets.}
Three datasets are used. (1)~\emph{Corridor}: a 410\,s teleoperated traverse
of a university building with polished floors and glass-walled conference rooms,
recorded with an Orbbec Femto Bolt (1280${\times}$720 RGB, 640${\times}$576
ToF depth) at ${\sim}30$\,FPS (12\,384 frames). A 459-frame high-invalidity segment dominated by worst-case reflective
surfaces is used for offline ablation. (2)~\emph{LILocBench}~\cite{lilocbench}: the
\texttt{dynamics\_0} sequence, a public benchmark recording of a building lobby
with moving pedestrians, replayed at 0.7$\times$ real-time for complete frame
evaluation (DA3-Small's 21\,FPS suffices for online costmap updates). (3)~\emph{NYU Depth V2}~\cite{nyudepthv2}: the standard 795-train /
290-validation labeled split for general-purpose student evaluation and teacher
distillation.

\subsection{Platform Deployment}\label{sec:platform}

Table~\ref{tab:perception-stack} reports four runtime configurations on the
Jetson platform. DA3-Small alone reaches 21.9\,FPS at 5.1\,GB. The full
event-triggered stack (DA3 + YOLOv8 + SAM2) reaches 21.8\,FPS within 6.5\,GB.
The compact student (P4) runs at 218\,FPS and 2.7\,GB, 10$\times$ faster than
DA3-Small at half the memory. A full-stack depth sensing system can therefore remain usable under a
constrained runtime budget, and a distilled student can assume the
depth-sensing role at a fraction of the cost when the environment is known.

\begin{table}[!t]
\centering
\caption{Perception stack configurations on Jetson Orin Nano. P1 = DA3 only;
P2 = DA3 + YOLOv8; P3 = full stack with event-triggered semantic refinement;
P4 = EfficientViT-B1 student (TRT FP16).}
\label{tab:perception-stack}
\setlength{\tabcolsep}{3pt}
\footnotesize
\begin{tabular}{lccccl}
\toprule
 & \textbf{FPS} & \textbf{Depth} & \textbf{YOLO} & \textbf{SAM} & \textbf{RAM} \\
\midrule
P1 (DA3)           & 21.9          & 47\,ms & --        & --        & 5.1\,GB \\
P2 (DA3+YOLOv8)    & 19.2          & 45\,ms & 23\,ms    & --        & 5.7\,GB \\
P3 (full, evt)     & \textbf{21.8} & 39\,ms & 18\,ms    & on-demand & 6.5\,GB \\
P4 (student TRT)   & \textbf{218}  & \textbf{4.6\,ms} & -- & --   & \textbf{2.7\,GB} \\
\bottomrule
\end{tabular}
\end{table}

End-to-end pipeline latency (camera callback through costmap update) adds
ROS serialization, fusion, and back-projection overhead; the DA3 path
(P1) sustains ${\sim}21$\,FPS end-to-end, while the student path (P4) is
limited by the camera frame rate (30\,FPS), not by inference. The
6.5\,GB ceiling of the full stack motivates the student study in
\S\ref{sec:r3}.

\subsection{Corridor Costmap Recovery}\label{sec:r1}

The corridor is the most direct test of the system: active-depth failure
is persistent here. We evaluate seven configurations that vary the obstacle
source and inflation strategy across
Nav2's \texttt{ObstacleLayer} and \texttt{InflationLayer}:
Base/A1 use sensor-only depth with fixed (0.09\,m) or
corridor-width inflation;
A2/A3 add DA3-Small learned depth with the same two
inflation strategies;
A4 uses DA3-only (no hardware depth) with corridor inflation;
A5/A6 use sensor+DA3 or DA3-only with class-aware inflation
(YOLOv8-driven per-class radii: person\,$>$\,glass\,$>$\,furniture\,$>$\,wall).

The offline ablation (Table~\ref{tab:corridor-offline}) shows that the additive
LiDAR+depth configurations preserve baseline detection rate while increasing
obstacle evidence. A2, A3, and A5 maintain 100\% detection while introducing
learned depth. The DA3-only variants (A4, A6) are weaker, consistent with the
system design; these serve as the depth-replacement baseline, confirming that
learned depth must complement LiDAR, not replace it, to be reliable.

\begin{table}[!t]
\centering
\caption{Offline corridor costmap ablation (459-frame high-invalidity segment,
mean $\pm$ std). IoU and FPR use the LiDAR-only baseline (Base) as
reference; that baseline is itself incomplete (78\% invalid pixels), so
structure added by learned depth in invalid regions registers as false
positive. See \S\ref{sec:r1} text for interpretation.}
\label{tab:corridor-offline}
\setlength{\tabcolsep}{2pt}
\scriptsize
\begin{tabular}{@{}lcccrcc@{}}
\toprule
 & \textbf{IoU} & \textbf{Det.\,\%}
  & \textbf{FPR\,\%} & \textbf{Clear.\,(m)}
  & \textbf{Infl.\,(m)} & \textbf{ms} \\
\midrule
Base & 1.00            & 100               & 0.0            & .249$\pm$.007 & .090           & 16$\pm$6 \\
A1   & 1.00            & 100               & 0.0            & .249$\pm$.007 & .178$\pm$.005 & 63$\pm$27 \\
A2   & .379$\pm$.106   & 100               & 5.2$\pm$2.7    & .171$\pm$.051 & .090           & 87$\pm$26 \\
A3   & .379$\pm$.106   & 100               & 5.2$\pm$2.7    & .171$\pm$.051 & .178$\pm$.005 & 134$\pm$42 \\
A4   & .279$\pm$.102   & 76.7$\pm$9.9      & 5.2$\pm$2.7    & .175$\pm$.059 & .165$\pm$.021 & 133$\pm$42 \\
A5   & .379$\pm$.106   & 100               & 5.2$\pm$2.7    & .171$\pm$.051 & .192$\pm$.003 & 206$\pm$61 \\
A6   & .279$\pm$.102   & 76.7$\pm$9.9      & 5.2$\pm$2.7    & .175$\pm$.059 & .197$\pm$.003 & 190$\pm$56 \\
\bottomrule
\end{tabular}
\end{table}

IoU and FPR are computed against the LiDAR-only sensor map as the reference
baseline. Because that baseline is itself incomplete (78\% dead pixels in this
segment), any
structure filled by learned depth in those invalid regions counts as a false
positive under this metric. The drop in IoU when depth is added is therefore
expected: the new costmap contains more obstacle cells than the reference, so
the union grows while the intersection stays similar. A lower IoU reflects
richer coverage, not degraded accuracy. The metrics characterize how much
obstacle evidence learned depth adds relative to an incomplete reference.

\begin{figure}[!t]
\centering
\paperfigure{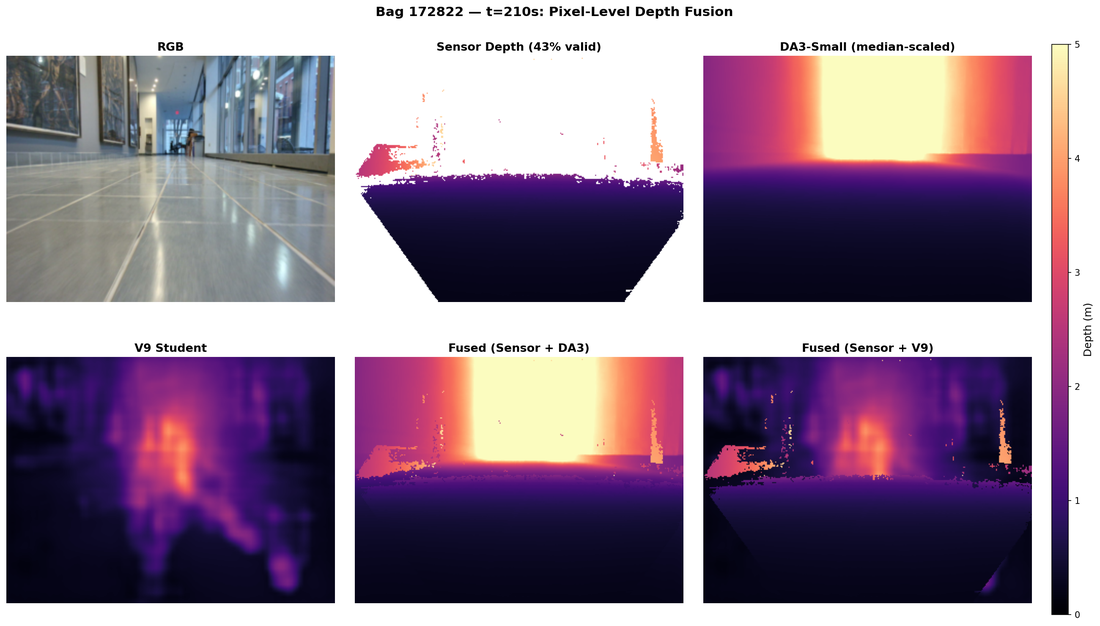}
{Pixel-level depth fusion at typical corridor failure (43\% valid pixels).}
\caption{Depth fusion on a representative corridor frame (43\% valid ToF
pixels). Top: RGB input showing polished floor and glass doors, sparse
sensor depth with large invalid regions, and DA3-Small depth prediction. Bottom: V9 corridor-specialist student, and both fused outputs
(sensor+DA3, sensor+V9). Learned depth fills the floor and glass regions
where the ToF sensor returns zero depth.}
\label{fig:depth-fusion-typical}
\end{figure}

Class-aware inflation (A5, A6) improves inflation precision through per-class
radii but does not improve detection rate or FPR relative to A2/A4. The
2.4$\times$ latency increase (206\,ms versus 87\,ms for A2) limits its use to
scenarios where per-class safety margins are required; it is not recommended as
a default configuration. When the student is active, the segmentation head
adds no inference cost (it shares the depth encoder), so per-class radii
can be enabled selectively for routes with frequent pedestrian traffic.

\begin{figure}[!t]
\centering
\paperfigure{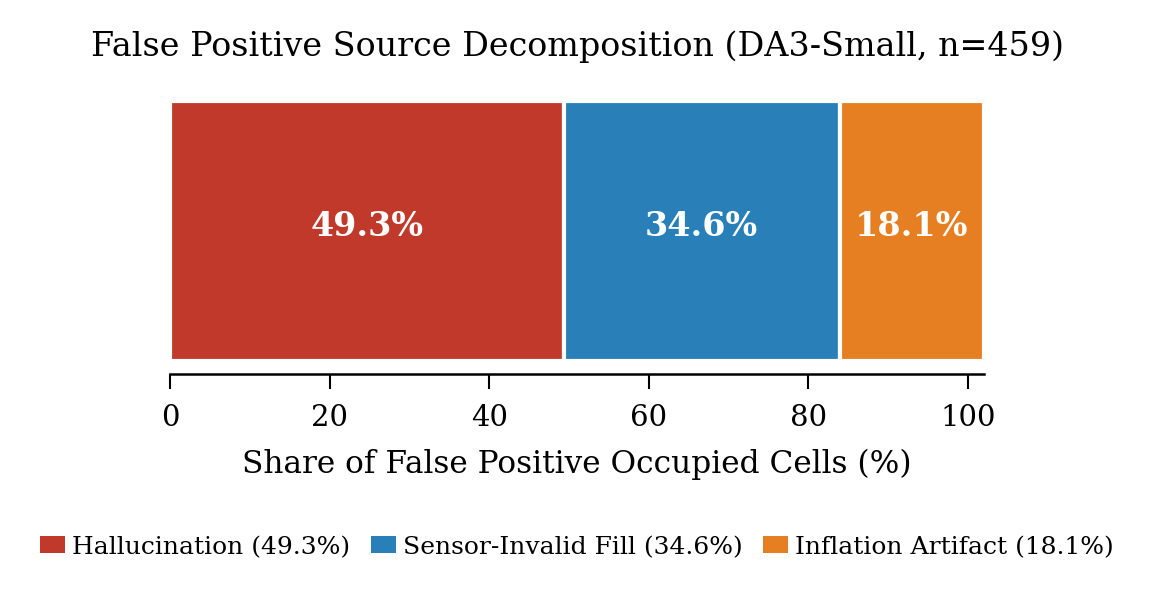}
{False positive source decomposition for DA3-Small (n=459).}
\caption{False-positive source decomposition (DA3-Small, $n{=}459$ corridor
frames). Sensor-invalid fill (34.6\%) represents structure added in ToF
dead-pixel regions; hallucination (49.3\%) is free-space false occupancy;
inflation artifacts (18.1\%) arise from Nav2 inflation of false cells.}
\label{fig:fpr-decomposition}
\end{figure}

The false-positive behavior is more informative than the raw 5.2\% suggests.
An audit separates that occupancy into three sources: 34.6\% arises in
sensor-invalid regions where learned depth fills the exact gaps that motivate
the system, 49.3\% corresponds to free-space hallucination by the learned
model, and 18.1\% comes from inflation artifacts
(Fig.~\ref{fig:fpr-decomposition}). The decomposition matters:
just under half of the flagged occupancy is genuine hallucination, not
legitimate gap-fill. The net benefit is increased obstacle coverage in
sensor-invalid regions; it is not uniformly improved geometric accuracy.

\begin{figure}[!t]
\centering
\paperfigure{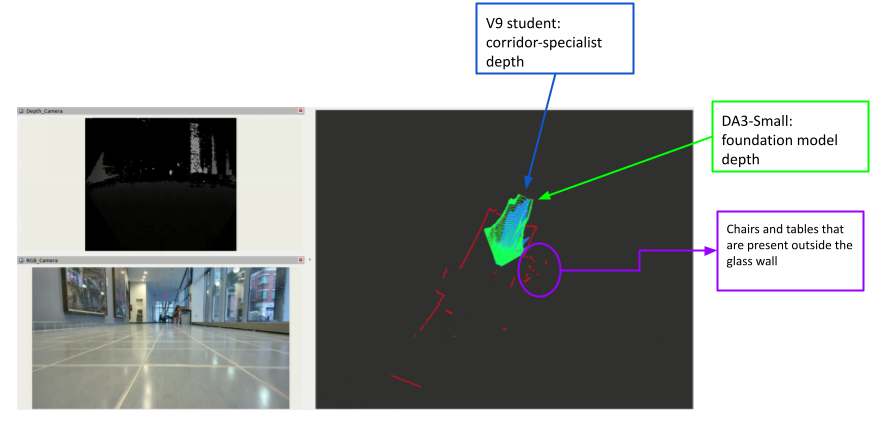}
{Live Nav2 costmap overlay in the corridor.}
\caption{Live Nav2 costmap during corridor replay. Left: hardware depth
(top, mostly invalid) and RGB camera (bottom). Right: costmap overlay
showing DA3-Small depth contributions (green) and V9 student depth (blue).
Learned depth detects chairs and tables visible through the glass wall
where the ToF sensor returns invalid depth.}
\label{fig:costmap-viz}
\end{figure}

The live replay result is shown in Table~\ref{tab:live-replay}. In the
corridor, LiDAR+depth increases occupied cells from 2295 to 3546, a 55\%
increase over LiDAR alone. LiDAR+sensor depth also helps, but less strongly,
because the active sensor is degraded in the same scene. Qualitatively, learned depth fills regions where the active sensor returns
effectively infinite depth, adding obstacle evidence for chairs, doorways, and
surfaces behind glass. Fig.~\ref{fig:depth-fusion-worstcase} shows the
worst case; Fig.~\ref{fig:depth-fusion-typical} shows typical failure rates.
Fig.~\ref{fig:costmap-viz} shows the resulting live Nav2 costmap.

Beyond coverage gains, the replay reveals an unexpected stability benefit.
Learned depth is often assumed to add frame-to-frame flicker, but here the
opposite holds: adding it \emph{reduces} normalized occupancy jitter from
0.61\% (LiDAR-only) to 0.43\% (LiDAR+depth) and centroid drift from
0.7\,mm to 0.5\,mm per frame across the 459-frame high-invalidity segment.
Why? The ToF sensor's dead-pixel pattern fluctuates between frames: pixels
that return valid depth on one frame may return zero on the next, causing
obstacle cells to appear and vanish. DA3-Small's temporally smooth dense
prediction stabilizes the regions where the ToF signal is intermittent.

\begin{table}[!tb]
\centering
\caption{Live costmap replay in the corridor and on LILocBench. L = LiDAR, S =
hardware depth, D = learned monocular depth, dyn = dynamic inflation.
Occ.\ = occupied cells (grid resolution 0.05\,m/cell).}
\label{tab:live-replay}
\setlength{\tabcolsep}{2.5pt}
\footnotesize
\begin{tabular}{@{}l rr@{\;\;}c rr@{\;\;}c@{}}
\toprule
 & \multicolumn{3}{c}{\textbf{Corridor}} & \multicolumn{3}{c}{\textbf{LILocBench}} \\
\cmidrule(lr){2-4}\cmidrule(lr){5-7}
\textbf{Config} & \textbf{Occ.\,(cells)} & \textbf{Std} & $\boldsymbol{\Delta}$
  & \textbf{Occ.\,(cells)} & \textbf{Std} & $\boldsymbol{\Delta}$ \\
\midrule
L       & 2295 & $\pm$312  & ---            & 957  & ---       & --- \\
L+S     & 3061 & $\pm$308  & +33\%          & 1106 & $\pm$285  & +16\% \\
L+D     & 3546 & $\pm$283  & \textbf{+55\%} & 2000 & $\pm$431  & \textbf{+110\%} \\
D$^\dag$  & 1461 & $\pm$159  & $-$36\%        & 3053 & $\pm$643  & +219\% \\
L+D+dyn & 3740 & $\pm$326  & +63\%          & 2115 & $\pm$368  & +122\% \\
\bottomrule
\end{tabular}
\par\noindent{\tiny $^\dag$D-only loses 36\% in corridor (no LiDAR anchor under ToF failure) but gains 219\% in LILocBench (healthy ToF, depth adds vertical extent). This asymmetry motivates the L+D hierarchy.}
\end{table}

\subsection{Dynamic Obstacle Recovery (LILocBench)}\label{sec:r2}

Here the problem is different: the 2D LiDAR is geometrically stable but
sees only a single horizontal slice through moving pedestrians. A person's
legs may register on the scan plane, but the torso, arms, and head (which
define the full collision envelope) are invisible to a 2D sensor mounted at
ankle height. Learned depth contributes the vertical extent that LiDAR
structurally cannot observe.

In Table~\ref{tab:live-replay}, LiDAR+depth increases occupied cells from
957 to 2000 (+110\% over LiDAR alone). Depth-only reaches 3053 (+219\%),
the highest absolute count in either dataset. This asymmetry with the
corridor results (where D-only drops 36\%) is expected: in LILocBench the
ToF sensor is not failing, so there is no reflective-surface penalty; the
depth model sees full pedestrian geometry that LiDAR simply cannot, while
in the corridor the depth model must compensate for active-sensor failure
with less reliable learned estimates. The combination L+D captures both
LiDAR's angular precision and depth's vertical coverage, which suggests that it
provides the most operationally useful costmap across the two datasets.
Although D-only reaches a higher raw cell count in LILocBench, it lacks the
geometric anchoring that LiDAR provides.

\begin{figure}[!t]
\centering
\paperfigure{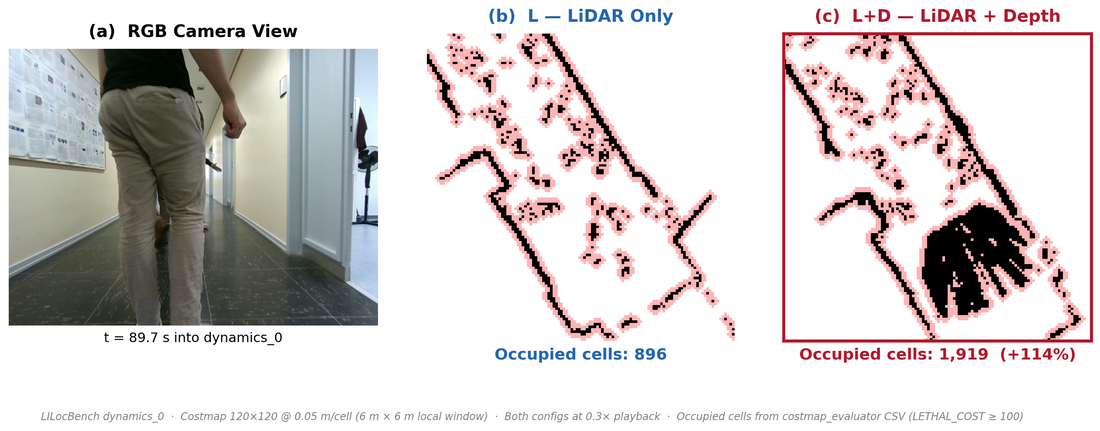}
{Dynamic benchmark. RGB + costmap comparison at t=89.7s: LiDAR-only versus
LiDAR+learned-depth in the pedestrian scene.}
\caption{LILocBench \texttt{dynamics\_0} at $t{=}89.7$\,s. (a)~RGB: pedestrian
at close range. (b)~LiDAR-only: 896 occupied cells (leg-height structure only).
(c)~LiDAR+depth: 1\,919 occupied cells (+114\%), capturing pedestrian
body extent above the 2D scan plane. Grid: $120{\times}120$ at 0.05\,m/cell
($6\,\text{m}{\times}6\,\text{m}$ local window); 0.7$\times$ playback to
match inference throughput.}
\label{fig:lilocbench-comparison}
\end{figure}

Fig.~\ref{fig:lilocbench-comparison} shows a representative frame.
Learned depth fills the pedestrian body above the scan plane, increasing
occupied cells from 896 to 1\,919 in a single frame. The additional cells
correspond to torso and shoulder structure that would otherwise be invisible
to the costmap.

\subsection{Compact Student for Navigation}\label{sec:r3}

The full DA3 stack consumes 6.5\,GB and runs at 21.8\,FPS, leaving little
headroom on an 8\,GB device. At 218\,FPS and 2.7\,GB, the compact student
frees over 3\,GB for the autonomy stack. Whether a model this small can
still produce depth good enough for navigation in a known environment is
the practical question.

Table~\ref{tab:distillation} reports a systematic ablation of six design
decisions organized in two phases. Phase~1 builds the general-purpose student
cumulatively: switching from MobileNetV3-Small to EfficientViT-B1 reduces NYU
RMSE from 1.160\,m to 0.774\,m; adding deployment augmentation brings it to
0.572\,m; adding SUN~\cite{sunrgbd}/DIODE~\cite{diode} pretraining reaches 0.510\,m. None of these general
improvements reduce LILocBench RMSE below 1.3\,m.

Phase~2 tests three domain-adaptation strategies from the best general
checkpoints. Mixed-domain training (NYU and corridor data combined) preserves
general performance (0.592\,m) but does not reduce LILocBench RMSE. Pure
corridor fine-tuning closes the domain gap, reaching 0.445\,m from the
augmented checkpoint and 0.382\,m from the pretrained checkpoint, but causes
catastrophic
forgetting (NYU RMSE degrades to 1.3--1.6\,m). The pretrained warm-start
produces the strongest corridor specialist, outperforming both the earlier
fine-tune and the foundation baseline (DA3-Small, 0.596\,m).

\begin{table}[!tb]
\centering
\caption{Student distillation ablation. Phase~1 builds the general-purpose
student cumulatively; Phase~2 tests domain-adaptation strategies from the best
Phase~1 checkpoints. LILoc RMSE is evaluated on LILocBench \texttt{dynamics\_0}.}
\label{tab:distillation}
\setlength{\tabcolsep}{2pt}
\scriptsize
\begin{tabular}{@{}lp{32mm}rrr@{}}
\toprule
 & \textbf{Design Decision} & \textbf{NYU RMSE}
  & \textbf{mIoU} & \textbf{LILoc RMSE} \\
\midrule
\multicolumn{5}{@{}l}{\emph{Phase~1: General-purpose student (cumulative)}} \\[2pt]
V3  & MNv3-S, berHu+CE, Kendall (baseline) & 1.160\,m & 39.3\% & --- \\
V4  & \enspace+ EfficientViT-B1 backbone & 0.774\,m & 51.0\% & 1.373\,m \\
V5  & \enspace+ deployment augmentation & 0.572\,m & 63.7\% & 2.186\,m \\
V6  & \enspace+ SUN/DIODE pretraining & 0.510\,m & 50.5\% & 2.158\,m \\[2pt]
\midrule
\multicolumn{5}{@{}l}{\emph{Phase~2: Domain adaptation strategies}} \\[2pt]
V7  & Corridor FT from V5 & 1.315\,m$^\dagger$ & 47.5\%$^\dagger$ & 0.445\,m$^\ddagger$ \\
V8  & Mixed NYU+corridor from V5 & 0.592\,m & 62.9\% & 2.266\,m \\
V9  & Corridor FT from V6 & 1.553\,m$^\dagger$ & 31.6\%$^\dagger$ & \textbf{0.382\,m}$^\ddagger$ \\[2pt]
\midrule
DA3-S & Foundation baseline (DA3-Small) & 0.513\,m$^\S$ & --- & 0.596\,m \\
\bottomrule
\end{tabular}
\vspace{1pt}
\par\noindent{\tiny
$^\dagger$After corridor FT; NYU performance degrades (catastrophic forgetting).
$^\ddagger$Evaluated on LILocBench \texttt{dynamics\_0} (RealSense D455); not directly comparable to Femto Bolt corridor data.
$^\S$290-image NYU validation with per-image median scaling.
MNv3-S = MobileNetV3-Small. All V4--V9 use EfficientViT-B1 (5.31M params).}
\end{table}

The ablation reveals a clear tradeoff: corridor specialization can beat
DA3-Small on LILocBench RMSE (0.382\,m versus 0.596\,m), but only by
sacrificing general performance. Mixed training (V8) does not bridge this
gap. The student therefore suits fixed routes in known buildings where short
offline retraining is acceptable and the deployment environment matches the
fine-tuning data. Outside that scope, DA3-Small remains the safer choice.

\subsubsection{Near-Range Safety Bound}

The 0.3--1.0\,m range is where depth errors translate directly into
collisions: an obstacle at 0.5\,m that the model places at 1.5\,m will not
enter the costmap's inflation radius.
Table~\ref{tab:nearrange} reports depth accuracy in this band on corridor data.
DA3-Small reaches 0.158\,m RMSE with $\delta < 1.25$ of 70.7\%, meaning
roughly 71\% of near-range pixels have less than 25\% depth error, the only
result here credible for autonomous near-range navigation. The general student
and the corridor specialist from the augmented checkpoint both achieve
$\delta < 1.25$ of effectively 0\%, placing almost no pixels within 25\% of
the correct depth. The strongest corridor specialist improves to $\delta <
1.25 = 8.83\%$ (RMSE = 1.642\,m), better than the earlier variants but still
far below the foundation model. This gap is the primary barrier to deploying
the student for autonomous navigation without a safety supervisor.

\begin{table}[!t]
\centering
\caption{Near-range depth accuracy in the 0.3--1.0\,m corridor band.}
\label{tab:nearrange}
\setlength{\tabcolsep}{3pt}
\footnotesize
\begin{tabular}{lrrrr}
\toprule
\textbf{Model} & \textbf{RMSE} & \textbf{MAE} & \textbf{AbsRel} & \textbf{$\delta < 1.25$} \\
\midrule
Foundation (DA3-Small)    & 0.158\,m & 0.087\,m & 0.169 & 70.7\% \\
General student (V4)     & 1.434\,m & 1.389\,m & 2.918 & 0.0\% \\
Corr.\ FT from V5 (V7)  & 1.982\,m & 1.499\,m & 2.882 & 0.0\% \\
Corr.\ FT from V6 (V9)  & 1.642\,m & 1.252\,m & 2.440 & 8.83\% \\
\bottomrule
\end{tabular}
\end{table}

Concretely, $\delta{<}1.25 = 8.83\%$ means that at 0.5\,m range, roughly
91\% of pixels have more than 25\% depth error: an obstacle at 0.5\,m
could be placed at 0.63\,m or beyond in the costmap, outside typical
inflation radii. The student can become a strong fixed-route specialist,
but near-range reliability requires a safety supervisor for general
deployment outside fixed routes.

\subsubsection{Closed-Loop Simulation}

We tested whether the corridor-specialized student (V9, TensorRT FP16)
produces depth sufficient for planner-level operation in an idealized
setting. A 12\,m narrow corridor (1.2\,m passable width) with walls on both
sides forms the simulated environment. Model Predictive Path Integral (MPPI) control drives the robot using
the student's depth-derived costmap as the sole obstacle source (no LiDAR
in this test). Gazebo provides synthetic
ground-truth depth free of reflective-surface failures; this provides an
upper-bound reference against the student.
Ten random seeds per depth source vary the initial pose noise
(Table~\ref{tab:gazebo-sim}).

\begin{table}[!t]
\centering
\caption{Closed-loop Gazebo simulation (narrow corridor, 10 seeds per profile).
TTG = time-to-goal.}
\label{tab:gazebo-sim}
\setlength{\tabcolsep}{3pt}
\footnotesize
\begin{tabular}{lcccr}
\toprule
\textbf{Depth Source} & \textbf{Success} & \textbf{TTG (s)} & \textbf{Collisions} & \textbf{Path (m)} \\
\midrule
Ground truth      & 9/10 & 17.8$\pm$0.9 & 0 & 12.20$\pm$0.01 \\
Corridor student (TRT)  & 9/10 & 18.0$\pm$0.4 & 0 & 12.20$\pm$0.01 \\
\bottomrule
\end{tabular}
\end{table}

The corridor student matches the ground-truth baseline: same success rate,
zero collisions, near-identical path length and time-to-goal (0.2\,s
difference). The single failure in each profile is a recovery timeout: MPPI
enters a local costmap minimum near the goal and the robot stops rather than
collides. This failure mode is identical for both depth sources, confirming
it is a planner limitation, not a depth-quality issue.

Despite its weaker near-range accuracy (Table~\ref{tab:nearrange}), the
student produces depth sufficient for collision-free corridor traversal in
this simulated setting. Gazebo lacks reflective surfaces, ToF failure, and
lighting variation, so the result bounds task feasibility rather than
real-world performance.

\section{Conclusion}

When a ToF camera loses most of its depth pixels, the surviving valid
measurements can still anchor learned monocular depth to metric scale.
We demonstrated this self-calibrating property for navigation costmap
construction: the degraded sensor both motivates and calibrates the learned
substitute at runtime, without offline data or external ground truth.
The sensing hierarchy built on it raises costmap obstacle coverage by 55\% in a reflective corridor and
110\% on a public pedestrian benchmark relative to LiDAR alone, without
discarding hardware depth where it remains reliable. A distilled corridor
specialist matches ground-truth navigation in simulation at 218\,FPS and
2.7\,GB, though its near-range accuracy limits unsupervised general
deployment.

Three findings emerge from these results. First, selective learned-depth
substitution, triggered by hardware failure instead of running always on, improves
costmap obstacle coverage relative to LiDAR alone in the tested settings.
Learned-depth fill also reduces costmap flicker caused by intermittent
ToF dead pixels. Second, the sensing hierarchy itself is the contribution:
failure-aware ordering preserves hardware depth where valid and fills only
where it fails, enabling conservative behavior under reliable sensing
and recovery under degraded sensing. Third, knowledge
distillation to a compact student is viable for fixed-route deployment, with
the deployment boundary set by near-range safety: DA3-Small
($\delta{<}1.25 = 70.7\%$, RMSE~0.158\,m) remains the recommended general
runtime; the corridor student ($\delta{<}1.25 = 8.83\%$, RMSE~1.642\,m) is
best viewed as a fixed-route specialist, suitable where the route is known,
retraining is acceptable, and an external safety layer is available.

Closed-loop hardware deployment and evaluation in industrial settings
(warehouse floors, manufacturing lines) where the same reflective-surface
failures occur remain future work.
Code and data will be made publicly available.

\balance
\bibliographystyle{IEEEtran}
\bibliography{references}

\end{document}